\documentclass[sigconf]{acmart}
\usepackage{booktabs} % For formal tables
\usepackage{natbib}

\usepackage{todonotes}
\usepackage[page]{appendix}

\definecolor{attngreen}{RGB}{53, 137, 77}
\setcopyright{none}
\renewcommand\footnotetextcopyrightpermission[1]{} % removes footnote with conference information in first column
\begin{document}
\title{Point-less: More Abstractive Summarization with Pointer-Generator Networks}
%\titlenote{Produces the permission block, and  copyright information}
\subtitle{}
%\subtitlenote{The full version of the author's guide is available as \texttt{acmart.pdf} document}

\author{Freek Boutkan}
\affiliation{%
  \institution{University of Amsterdam}
}

\author{Jorn Ranzijn}
\affiliation{%
  \institution{University of Amsterdam}
}

\author{David Rau}
\affiliation{%
  \institution{University of Amsterdam}
}

\author{Eelco van der Wel}
\affiliation{%
  \institution{University of Amsterdam}\vspace{2cm}%this is a hack but it works
}

\begin{abstract}
The Pointer-Generator architecture has shown to be a big improvement for abstractive summarization seq2seq models. However, the summaries produced by this model are largely extractive as over 30\% of the generated sentences are copied from the source text. This work proposes a multihead attention mechanism, pointer dropout and two new loss functions to promote more abstractive summaries while maintaining similar ROUGE scores. Both the multihead attention ,and dropout do not improve N-gram novelty, however, the dropout acts as a regularizer which improves the ROUGE score. The new loss function achieves significantly higher novel N-grams and sentences, at the cost of a slightly lower ROUGE score.
\end{abstract}

% \keywords{Abstractive Summarization}

\maketitle

\section{Introduction}
 More data is becoming available on the web every day, for instance in the form of news articles and scientific publications, and extracting the most relevant information is becoming increasingly difficult. A well-written summary should be able to provide the gist of a text and can help to reduce the effort of obtaining all the relevant information. 
 Automated text summarization has therefore received a lot of interest since the advent of deep learning techniques \citep{allahyari2017text}. The process of summarization is often divided into extractive summarization and abstractive summarization. In extractive summarization, a summary is obtained by copying the relevant parts of the text. Abstractive summarization aims to distil the relevant information from the source text into a summary by paraphrasing and is therefore not limited to the use of the exact phrases of the source text. 
 
\citet{see2017get} propose a new approach for abstractive summarization by combining a sequence-to-sequence model \cite{sutskever2014sequence} with a Pointer Network \cite{vinyals2015pointer}.  Additionally, See et al. \cite{see2017get} incorporate a coverage mechanism that aims to tackle the problem of over-generation by penalizing attention to words in the source document that have already received attention in past timesteps. 
 
 One issue with pointer-generator networks is that, during test time, the model focuses mainly on the source text during summary generation and does not introduce many novel words. The resulting summaries therefore tend to be more extractive than abstractive \cite{baumel2018query, see2017get}. \citet{see2017get} state that this focus on the source text is likely caused by the word-by-word supervision during training time, which is not possible during test time. Another issue with the pointer-generator architecture is that the generator is undertrained as the network learns to use the pointer mechanism early in training and arrives at a local minimum. We hypothesise that these two factors are the main contributors to the over-reliance on the pointer mechanism during test time.

 Another limitation is the used dataset and evaluation metric ROUGE. This is discussed in detail in section \ref{sec:relwork}. In abstractive summarization there are many candidate solutions. However, the provided dataset rarely contains all of these and perfectly viable summaries are sometimes penalized. This is closely related to the problem with the ROUGE metric as this can also produce low score for viable summaries \cite{see2017get}. These problems are not specific to the pointer-generator model, and addressing them is less obvious. %TODO: back this up with papers
 
%  \citet{kryscinski2018improving} improve the abstraction in text summarization by directly optimizing the ROUGE metric, however by making use of reinforcement learning. % Seems a bit out of place here? Do we want to criticize this approach?
 
The goal of this research to increase the number of novel N-grams while obtaining similar ROUGE scores, therefore improving abstraction in an end-to-end trainable text summarization model. Our contributions comprise of
 \begin{itemize}
    %  \item scheduled teacher forcing to anneal train and test conditions
     \item Multihead attention over the source text
     \item Dropout mechanism over the pointer
     \item Naive pointer regularization
     \item Pointer regularization based on word priors
 \end{itemize}

% \section{Ideas}
% Main issue: Model favours extractive (pointer) over abstractive (generator)

% \begin{itemize}
%     \item Regulate/penalize $p_{gen}$, the factor that decides if the model output uses the pointer or generator. \citet{weber2018controlling} have a similar idea by adding a penalty term to the beam-search. Other ideas: penalize consecutive pointers, so copying whole phrases is less likely. Use high dropout/regularize $p_{gen}$ to enforce sparsity so network can rely less on pointer.
%     \item Paper uses a single distribution for attention and pointing. Multiple attention distributions (multi-head attention).
%     \item Modify teacher forcing to something like professor forcing \citet{lamb2016professor}. 
% \end{itemize}

\section{Related Work}
\label{sec:relwork}
% Cite some cool related work and say something smart
% The main issue that we are trying to solve of the original pointer-generator network for abstractive summarization is the overreliance on the pointer network during test time. 

\subsubsection*{Abstractive and extractive summarization}

In extractive summarization, fragments of the source text are concatenated to generate a summary \cite{allahyari2017text}. An advantage of this task is that it is relatively easy to obtain a summary with good fluency and factual correctness. 
In contrast, abstractive methods allow for the use of synonyms, generalization, and rephrasing of the source text. While in theory this can lead to results that are closer to human generated summaries \citet{jing2002using}, it is a much more difficult task than extractive summarization. Common problems include factual and grammatical mistakes but also over/under generation of words \cite{allahyari2017text}.

In more recent work \cite{gu2016incorporating, see2017get}, hybrid models are proposed to combine the strengths of both methods. These models can create abstractive summaries with extractive elements to promote factual correctness, and out-of-vocabulary (OOV) word generation.

% Extractive = select sentence (fragments) \citet{allahyari2017text}. Relativly easy. fluency. not likely to introduce mistakes. 
% Abstractive = rewrite. Allows for synonyms, generalization, rephrasing. More difficult. grammatical sentences. Can introduce factual mistakes. Often LSTM and suffers common problems (under/over generation, no OOV generation). Human summaries are more abstractive \citet{jing2002using}
% Hybrid models: combine both. Some abstractive models rely on extractive preprocessing \citep{allahyari2017text}.
\subsubsection*{Pointer networks}
The incorporation of a copying mechanism to the sequence-to-sequence has proved to be a powerful addition for summarization tasks. Both the CopyNet \cite{gu2016incorporating} and Pointer-Generator \cite{see2017get} propose adding such a mechanism to bypass the generator network, in order to generate words directly from the input document. While this is useful in many cases, both papers observe balancing the strength of the pointer mechanism and the generator is a difficult task. The pointer generator seems easier to train and, as a result, most of the generated summary is generated by directly copying from the source.

\citet{weber2018controlling} confirmed the over-reliance on the pointer mechanism and introduced a penalty during beam decoding in order to increase the probability of generating a word from the generator distribution. However, no changes are made to the training process and the clear downside of this approach is that the over-reliance is not solved during training time but only afterwards.

\citet{song2018structure} add structural elements to the copy mechanism. They say a possible problem of the copy mechanism is that it only looks at semantic information, while structural information (such as grammatical structure) might be more important for generating good summaries.

% \cite{li2017deep} make the traditional sequence-to-sequence abstractive summarization more powerful with by adding a Variational Auto-encoder subnetwork to the decoder. While a more powerful decoder might be able to produce more abstractive summarizations, it is not clear why the addition of a generative component is needed.

\subsubsection*{Attention}
A main difference between the Copynet architecture \cite{gu2016incorporating} and Pointer-Generator \cite{see2017get} is that Copynet uses a separate attention distribution for pointing and generating, while the Pointer-Generator only uses one. \citet{see2017get} pose that similar information is needed for both pointing and generating, and that decoupling the two distributions might lead to a loss in performance. However, a popular recent architecture proposed for machine translation takes an opposite approach. %bruggetje!
\citet{vaswani2017attention} propose a multi-head attention mechanism, which is able to learn multiple attention distribution over an input sequence. These attention mechanisms are merged and projected with a linear layer, and can theoretically encode a more varied representation of the input sequence compared to the regular attention mechanism.

\citet{fan2018robust} are the first to use multi-head attention with a comparable model architecture for abstractive summarization. They show that multi-head mechanisms are useful for summarization tasks and that different useful features are learnt by the different attention heads. This could be particularly useful for the pointer-generator, since the distribution used by the pointer, and the distribution for the generator are identical in the original architecture.

\subsubsection*{Model evaluation}
\label{sec:eval}
Generated summaries will be compared against provided target summaries.
The \textit{ROUGE} score \citet{rouge} indicates the recall of overlapping N-grams between the generated and target summary.
Using ROUGE as the evaluation metric is problematic as has been noted by \citet{dohare2017text}.
Not only because ROUGE scores do not correlate with human judgement but more fundamentally because ROUGE can not evaluate restructured sentences in a proper way.
ROUGE matches overlap in complete words and in reconstructed sentences different word forms can be used which might lead to low ROUGE scores.
\citet{see2017get} shows an example of a valid summary that has a ROUGE score of 0.

\citet{krantz2018abstractive} propose a new metric \textit{VERT} that compares similarity scores of sentences.
Since this method does not match exact word forms and it is to some extend robust to grammatical changes, word reordering and sentence reconstructions.
Versatile Evaluation of Reduced Texts (VERT) is made up out of a similarity and dissimilarity sub-score. A sentence vector is created out of the reference and created summaries and the cosine similarity between these two vectors is measure of semantic similarity between the summaries. The dissimilarity sub-score is calculated using the word-mover-distance algorithm that indicates how much a created summary has to change in order to match a reference summary. This new metric correlates stronger with human judgement compared to the commonly used ROUGE metric.

Both ROUGE and VERT only measure the accuracy of the generated sentences with respect to the target summary but they provide no insight in the \textit{abstractiveness}.
To measure abstractiveness we use the proportion of new N-grams in the generated summary.
A low proportion of higher order N-grams indicates that the model is copying long phrases from the input sequence and is thus acting in a more extractive way.
Improving both ROUGE and novel N-grams seems like a contradiction since improving ROUGE will decrease the number of new N-grams if the generated summary is not rephrased in the same way as the reference summary.  %But for ROUGE only lower order N-grams will be used whereas for abstractiveness we will mainly look at higher order N-grams. 

\subsubsection*{Directly improving novel N-grams using policy learning}
\citet{kryscinski2018improving} optimize the ROUGE score directly. Since the ROUGE metric is not differentiable this can only be done by using reinforcement techniques such as policy improvement. The loss function combines the maximum likelihood and ROUGE objective. In addition an abstractive reward is added to the loss. This reward is defined as the proportion of novel N-grams in the generated summary. This metric has a bias towards very short summaries and needs to be normalised using the length ratio of the generated and  ground-truth summaries. They achieve similar ROUGE scores as \cite{see2017get} but show that the number of new N-grams increases significantly and thus is less extractive.
% Might not be needed?

\section{methods}
In this section we describe our dataset (\ref{sec:dataset}) and (\ref{sec:baseline}) the baseline pointer-generator network.  Then we introduce our extensions over the baseline network which comprises our multi-head attention (\ref{sec:multihead}), pointing penalty losses (\ref{sec:pointing_losses}) and pointer dropout mechanism (\ref{sec:dropout}). 

\subsection{Dataset}
\label{sec:dataset}
We use the CNN Dailymail dataset \citet{hermann2015teaching}. We use the same preprocessing and training splits as \citet{see2017get}, which in turn uses the method from \citet{nallapati2016abstractive}. The training set consists of approximately 287k training pairs, with a validation set of 13 thousand pairs and test set of 11 thousand examples. The average article length is 781 tokens and the summary length is on average 56 tokens ($\pm 3.75$ sentences). 

\subsection{Pointer-Generator network}
\label{sec:baseline}
The baseline model is the pointer-generator network described by \citet{see2017get}. This model allows for copying of words from the source document using a pointing mechanism and also generation of novel words by selecting words from a fixed vocabulary. The main advantage of this approach over previous methods is that it allows the model to produce out of vocabulary words during summary generation.

The basic architecture is a sequence-to-sequence attention model. Words from the source document are fed sequentially into a single bidirectional LSTM resulting in a sequence of encoder hidden states. The decoder is a single layer LSTM that is initialised with the final hidden states of the encoder. More specifically, a linear layer maps the final bidirectional hidden states to a fixed size output that represent the initial values of the decoder at the first time step. 
During the decoding at time step $t$, an attention distribution is calculated over the source words:
\begin{equation*}
    e^{t}_{i} = v^{T}\text{tanh}(W_{h}h_{i} + W_{s}s_{(t)} + W_{c}c_{i}^{t} + b_{att})
\end{equation*}
\begin{equation*}
a^{t} = \text{softmax}(e^{(t)})
\end{equation*}
Here, $v^{T}, W_{h}, W_{s}, W_{c}, b_{att}$ are learnable parameters. $s_{t}$ refers to the output of the decoder at time step $t$ and $h_{i}$ is the representation of the word at position $i$ produced by the encoder.
$c^{t}_{i} $ is the coverage vector.

\citet{see2017get} include a coverage mechanism in their model to reduce the amount of repetition. By taking into account the amount of attention that has been given to words from the source text in previous time steps, they manage to significantly decrease repetition in the produced summaries. The coverage vector at time step $t$ is just the sum of attention of the previous time steps:
\begin{equation*}
    c^{t} = \sum_{t'=0}^{t-1}a^{(t')}
\end{equation*}

The attention distribution indicates which words from the source text are relevant to produce the next word of the summary. This information is stored in a fixed size representation called the context vector, $h^{*}_{t}$, that is a weighted combination of the encoder hidden states:

\begin{equation*}
    h^{*}_{(t)} = \sum_{i}a_{i}^{(t)}h_{i}
\end{equation*}
Based on the context vector and decoder hidden state, a probability distribution is calculated over the fixed size vocabulary. The context vector and decoder hidden state are concatenated and subsequently fed through two linear layers and a softmax to obtain a valid probability distribution over the vocabulary words, called $p_{vocab}$.

\begin{equation*}
    p_{vocab} = \text{softmax}(V'(V[s_{(t)}, h^{*}_{(t)}] + b) + b')
\end{equation*}

Here, $V', V, b'$ and $b$ are the learnable parameters.
A copy distribution over the source words is also required in order to select words from the source text during summary generation. \citet{see2017get} decided to recycle the corresponding attention distribution and also made it serve as the pointing distribution. The probabilities of the words that occurred multiple times in the source text were summed.

A trade-off has to be made between copying a word with the help of the attention distribution and generating a word by the $p_{vocab}$ distribution. Therefore, a generation probability, $p_{gen}$ was introduced that acts as a soft switch. A $p_{gen}$ of 1 would mean that only words from the $p_{vocab}$ distribution can be used and none from the pointing distribution while a $p_{gen}$ of 0 has the opposite effect.

\begin{equation*}
    p_{gen}^{(t)} = \sigma(w^{T}_{h^{*}}h^{*}_{(t)} + w^{T}_{s}s_{(t)} + w^{T}_{x}x_{(t)} + b_{ptr})
\end{equation*}
\\
Here, $w^{T}_{h^{*}}, w^{T}_{s}, w^{T}_{x}, b_{ptr}$ are learnable parameters. $x_{t}$ refers to the input of the decoder at time step $t$ and $\sigma$ is the sigmoid function.

For every document, there is an extended vocabulary that is the union of the words in that document and all the words in fixed vocabulary. Now, the probability of a word in this extended vocabulary is defined as (where $p_{point} = 1 - p_{gen}$):

\begin{equation*}
    p(w) = p_{gen}
    \cdot p_{vocab}(w) + (p_{point}) \cdot \sum_{i:w_{i}=w}a^{(t)}_{i}
\end{equation*}

The loss function at time step $t$ is defined as:
\begin{equation*}
    loss_{t} = -\text{log}p(w^{*}_{(t)}) + \lambda \sum_{i}\text{min}(a^{(t)}_{i}, c^{(t)}_{i})
\end{equation*}
where the first term is the negative log likelihood of target word $w^{*}$ and the second term is the coverage loss. This coverage loss is introduced to penalize repeated attention to the same words and is reweighted by a hyperparameter $\lambda$. 
The final loss function is defined as the average loss over all time steps:
\begin{equation*}
    loss = \frac{1}{T} \sum^{T}_{t=0} loss^{(t)}
\end{equation*}

\subsection{Dropout mechanism}
\label{sec:dropout}
We propose a dropout mechanism on the pointer network to make the model less dependent on the pointer mechanism.
Generally, dropout is a simple method to prevent overfitting in neural networks \cite{srivastava2014dropout}, by dropping parts of the network during training. With a predefined probability weights are set to zero during training. This ensures that the model can not rely on hidden co-dependencies and generalises better.
During evaluation the pointer-generator model tends to rely to much on the pointer mechanism. The contribution of the generator network to the final output probability is on average only 17\%.
Our pointer dropout method can be implemented by randomly, setting $p_{gen}$ with probability $0.2$ to 1 during training, where a value of 1 makes the output distribution of the model the same as the output of the generator.
We expect the model to rely less on the pointing mechanism and use the copy mechanism only when necessary. Hopefully, this would result in a model that generates more abstractive summaries.

\subsection{Multihead attention}
\label{sec:multihead}
In the original paper, the pointer and the generator make use of the exact same attention distribution. In our opinion this is problematic because pointer and generator carry out different functions that require different underlying features. For example, the generator might use syntactical features to create a correct sentence structure or point to multiple words to create a more abstract summary. In contrast, the pointer only attends to words that it wants to copy to the summary. 

\begin{figure}
    \centering
    \includegraphics[width=0.8\linewidth]{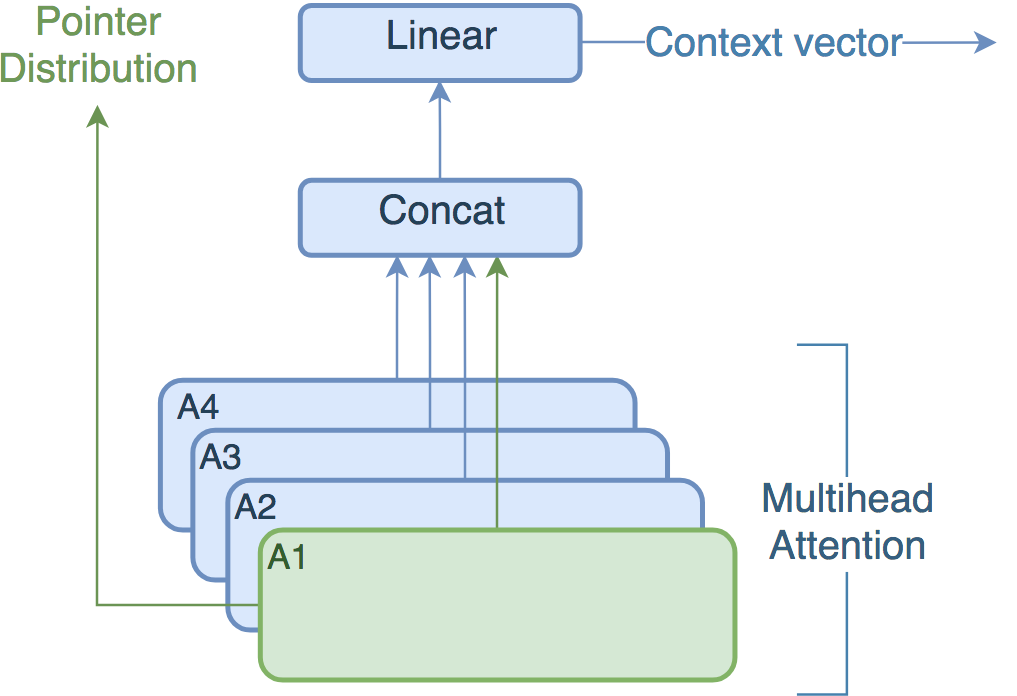}
    \caption{A schematic of the pointer-generator multi-head attention with four attention heads, where the first head is used as shared attention.}
    \label{fig:multihead}
\end{figure}

In order to both differentiate between pointer and generator attention distributions, but still supply all information of the pointer mechanism to the generator, we use a modification of the multi-head attention mechanism \cite{vaswani2017attention}. Figure \ref{fig:multihead} shows a schematic of our new pointer-generator multi-head attention mechanism where the {\color{attngreen}first attention head} is shared between the pointer and the generator, whereas the generator receives \textcolor{blue}{all attention heads}. This way, by introducing regularizations to the pointer mechanism, we only affect the shared attention head while dedicating the rest of the attention heads to generator specific features.
 
%  As our goal is to make a more abstractive summary we propose changes should affect only the pointer distribution without changing the generator.

% Additionally, when penalizing the pointer mechanism with different loss functions and dropout training to yield a more abstractive summarization, it can directly affect the attention distribuion. 

\subsection{Pointing losses}
\label{sec:pointing_losses}
\citet{see2017get} show that the model exploits the pointing mechanism during evaluation.
We hypothesise this is because pointing is easier than generating sentences.
The model takes a shortcut by copying a lot of phrases and sentences in cases where this might not be necessary.
To discourage the network to count on the pointer network we add a term to the loss called the naive pointing loss.
We add the sum of all pointing probabilities and weigh it with a hyperparameter $\lambda_p$.
This way the model can still use the pointer network but it will directly contribute to a higher loss. For readability, we define the pointer mechanism weight $p_{point} = 1 - p_{gen}$.
\begin{equation*}
\mathcal{L}_{naive} = \lambda_{p} \sum_{t=0}^T\left(p_{point}^{(t)}\right)
\end{equation*}

A disadvantage of this relatively simple penalty term could be that every word gets the same penalty, even for words where pointing is desired. We propose a second pointer penalty term, the word prior pointing loss, that only penalizes the pointer when a word is common in the vocabulary.

\begin{equation*}
    \mathcal{L}_{WP}^{(t)} = \lambda_p \cdot p_{point}^{(t)} \cdot - \sum_{w \in X} p_{w}(w) \log (1 - p_{a}^{(t)}(w))
\end{equation*}

Where $p_{a}^{(t)}$ is the attention distribution used for the pointer mechanism and $p_{w}$ is a pre-calculated word prior. As with the previous loss, $\lambda_p$ is a hyper parameter to weigh the influence of the loss during training.

Intuitively, the cross entropy between the prior $p_w$ and $(1-p_a^{(t)})$ expresses the surprisal of not pointing to a word given the word prior. If the prior is high, and the word has a high weight in the attention distribution, this loss term will be high. In any other case, the loss term will be small.

The desired behaviour of this loss term is that $p_{point}$ gets minimized when attending to words with a high prior. However, when naively implemented, the model could also minimize the cross entropy when $p_{point}$ is high. This would cause the model to attend to uncommon words while pointing, which is not what the loss is supposed to achieve. To prevent this, the gradient of this loss term is only back-propagated to $p_{point}$ during training.

\section{experiments}
% 1) Run baseline (COPY=True, Coverage=True)
% 2) Run TF decay experiment
% 3) Run Multihead attention
% If 2 + 3 do not make things dramatically worse 
% 4) Run Dropout with TF decay + Multihead
% 5) Run CrossEntropyloss with TF decay + Multihead
% 6) Run p_point loss with TF decay + Multihead

% Training the baseline model for 30 epochs as described by \citet{see2017get} takes approximately 4 days.

% Add that we only uses losses from the end as with coverage.
% Add that we train dropout from the start with 0.2 and experimented with different values but did not observe large differences.

This section describes the experimental setup (4.1) that is shared between all of the models and also the different model variations (4.2) that are tested.

\subsection{Experimental setup}
All the experiments follow the same setup as described by \citet{see2017get}. A fixed size vocabulary of 50.000 words is used for both the source and target words. All models use 128-dimensional word embeddings that are learned during training time and hidden states for the pointer and generator are kept at a fixed size of 256. The input summaries are truncated to 400 tokens during training and test time. The reason for this is that the most important words for the summary appear in the beginning of the articles and keeping longer source document even decreases performance \citep{see2017get}. Summary lengths are limited to 120 tokens during training time and 100 during test time in order to speed up training. Summaries are generated using beam search and use a beam size of 4 at test time.
All the model parameters are optimised with Adagrad \citep{Duchi:2011:ASM:1953048.2021068} using a learning rate of 0.15 and accumulator value of 0.1 as this proved to work best for \citet{see2017get}. Gradients are clipped with a maximum norm value of 2 and no further regularisation methods are used.

\subsection{Model variations}
The baseline model is the pointer-generator network described by \citet{see2017get}. The baseline is trained with and without coverage where the coverage mechanism is trained separately after 13 epochs for 3000 iterations as including this from the beginning turns out to decrease performance.

% Scheduled teacher forcing is only applied after the first 200.000 iterations. The reason for this is that the models should first perform sufficiently well before penalising undesired behaviour. Applying scheduled teacher from the beginning of training could be too severe and therefore diminish overall performance. % iets over inverse sigmoid decay?

The multi-head attention mechanism is tested with four heads. Every head produces a context vector that is $\frac{1}{4}$th of the size of the context vector of the baseline. Next, these context vector are concatenated into a single vector resulting in a vector of the same size that is independent on the number of heads. This is similar to the approach taken by \citet{vaswani2017attention}.

The probability for dropping out the pointer mechanism is set to 0.2. The decision to drop out the pointer holds for all the words during summary generation. This means that for some summaries, the model can not rely on the pointing mechanism at all. 

The pointing losses are added at the end of training and for the same amount of iterations as the coverage loss. For both losses experiments were conducted with different scalars (1, 0.4, 0.2, 0.05). As best performing scalars 0.05 for NLoss and 0.2 for WPLoss were found. 
Prior probabilities of the words are calculated on the occurrence in the entire training set. Also, only the prior probabilities for words that occur in the generation vocab are calculated. For the other words, this probability is set to zero.
Every adaptation to the baseline model that is proposed in this work is tested as a separate addition to the baseline. This gives a clear estimation of the influence of each adaptation although it leaves out the influence of possible interactions that might occur.
Due to the time intensive training we use the same hyperparameters for all models.

In order to test our results for \textbf{statistical significance} we perform Wilcoxon signed-rank tests \citet{wilcoxon1945individual}. All models without coverage are compared to the baseline whereas models with coverage are tested against the baseline + coverage model. Statistical tests for ROUGE 1, ROUGE 2 and ROUGE L are conducted. A \textit{p-value} of $0.01$ is used to determine statistical significance. Since the VERT score correlates strongly with the ROUGE scores separate tests are not needed. The differences in novel N-grams are generally much bigger and have lower variance than ROUGE scores so statistical tests are not needed for a large test set (11k examples).

\section{results}

\begin{table}[!h]
\caption{Mean ROUGE $\boldsymbol{F_{1}}$ and VERT scores of the tested models (11k examples in testset). All models were trained from epoch 13 on with coverage for 3000 steps. Here NLoss corresponds to naive pointing loss and WPLoss to the word prior pointing loss. Scores with a star are not significantly different from the baseline. Best results are marked in \textbf{bold}.}
%No need to mention training details
\resizebox{\columnwidth}{!}{%
\begin{tabular}{lllllll}
\toprule
%Model                 & Rouge 1 & Rouge 2 & Rouge L & VERT  \\ 
%Model & & & & & & & & \\
Attention heads & \multicolumn{2}{l}{Model extensions} & ROUGE 1 & ROUGE 2 & ROUGE L & VERT  \\ 
%MH & cov & Dropout & NLoss & WPLoss
\midrule
1 (\textit{baseline}) &   &              & 38.14   & 15.82   & 33.47   & 0.706   \\ % freek
 4&    &           & 38.15*   & 15.76*   & 33.47*   & 0.707 \\ % david
\specialrule{0.01pt}{1pt}{0pt}
 1&  dropout &  & \textbf{38.35} & \textbf{15.94} & \textbf{33.62}& \textbf{0.708} \\% david
 4& dropout  &    &38.09* & 15.75*&33.51* &0.706 \\% david
 \specialrule{0.01pt}{1pt}{0pt}
 %WPLoss (.05) + cov & 37.47   & 15.61   & 32.79   & 0.699\\ % freek
 %MH + WPLoss (.05) + cov  & 37.57   &  15.52  & 33.02   & 0.701  \\ % david
 %MH + WPLoss (.05) + cov  ( 2150 steps)& 36.73   &  15.17  & 31.87   & 0.694  \\ % david
 %WPLoss (.10) + cov        & 36.95       & 15.25        & 32.16      & 0.695     \\ % jorn 0.1
 %MH + WPLoss (.10) + cov  & 37.37   &  15.41  & 32.75   & 0.698  \\ % david
 1&    &  NLoss &  37.63 & 15.64 & 33.02 & 0.702 \\
 4&  & NLoss   &  37.35  & 15.41   &  32.74 & 0.699 \\
\specialrule{0.01pt}{1pt}{0pt}
 1& &   WPLoss  & 36.48   & 14.93  & 31.55   & 0.690 \\ % freek
4 &   &   WPLoss  & 36.81   &  14.95  & 32.13   & 0.693  \\ % david
 %MH + WPLoss (.10) + cov (7408 steps) & 36.74   &    15.14     &    32.04     &    0.694   \\  % david
 %2H + WPLoss(.2) + cov & 36.83 & 14.99 & 32.12 & 0.691 \\
 %2H + WPLoss(.05) + cov & 37.031 & 14.91 & 32.627 & 0.697 \\
\specialrule{0.8pt}{1pt}{0pt}
\multicolumn{2}{l}{1 \textit{(baseline \citet{see2017get}})} & & 39.53& 17.28& 36.38 & -\\
\bottomrule
\end{tabular}
}%
\label{model_results}
\end{table}

\begin{table}[]
\caption{Percentage of novel N-grams and sentences that are produced for each of the tested models. Best results are marked in \textbf{bold}.}
\resizebox{\columnwidth}{!}{%
\begin{tabular}{lllllllll}
\toprule
Attention heads & \multicolumn{2}{l}{Model extensions}         & 1-grams & 2-grams & 3-grams & 4-grams & Sentences \\
\midrule
1 (\textit{baseline})& &          & 0.17    & 3.24    & 8.12    & 12.80    & 79.52      \\ 
4 & &        & 0.12    & 2.90    & 7.56    & 12.12   & 78.38      \\
\specialrule{0.01pt}{1pt}{0pt}
1 & Dropout& &  0.12    & 2.67    & 7.13    & 11.51   & 75.94    \\
4 & Dropout& &  0.23  & 3.20    & 8.07   & 12.77   & 78.30      \\ 
\specialrule{0.01pt}{1pt}{0pt}
%WPLoss (.05) + cov     &   0.36      & 4.92     & 11.25   & 17.05   & 85.62                   \\ 
%WPLoss (.05) + MH + cov (2150 steps) &  0.39   &  5.26   &    11.92     &     17.95    &    88.26       \\
%WPLoss (.05) + MH + cov  &  0.23   &  4.64   &    11.14     &     17.15    &    84.72       \\
%WPLoss (.10) + cov       & 0.64    &  6.08   &    13.38     &     19.89    &    88.05       \\ 
%WPLoss (.10) + MH + cov  &  0.37   &  5.58   &    13.11     &     19.86    &    87.13       \\
1 & &NLoss    & 0.31 & 5.21 & 12.04 & 18.67 & 86.57\\
4 & &NLoss    & 0.30  & 5.06   & 11.92  & 18.17 & 86.20\\
\specialrule{0.01pt}{1pt}{0pt}
1 & &WPLoss    &     \textbf{1.44}    &   \textbf{8.86}      & \textbf{17.96}        & 25.50   &   \textbf{92.00}   \\ 
4 & &WPLoss    & 0.95     &     8.24    &     \textbf{17.96}    &      \textbf{26.07}   &        91.05   \\
% WPLoss (.10) + MH + cov (7408 steps) &    0.65     &     6.53    &     14.40    &      21.33   &        88.13   \\
% 2H + WPLoss(.20) + cov & 0.55 & 7.93 & 16.94 & 25.92 & 90.94 \\
% 2H + WPLoss(.05) + cov & 0.41 & 5.01 & 11.41 & 17.64 & 84.79 \\
\specialrule{0.8pt}{1pt}{0pt}
\multicolumn{3}{l}{Target summaries} & 16.95 & 52.48 & 72.36    & 81.94 & 98.97 \\
\bottomrule
\end{tabular}
}%
\label{tab:novelty}
\end{table}

In Table \ref{model_results} the average ROUGE scores are reported on all models, and Table \ref{tab:novelty} shows the amount of novel N-grams and sentences. These models are all trained with coverage, results without coverage are included in Appendix \ref{sec:appA}.

The obtained baseline ROUGE scores are on average 3 points lower than reported in \citet{see2017get} paper. However, we use a pytorch re-implementation \footnote{\url{https://github.com/lipiji/neural-summ-cnndm-pytorch}} and did not perform any hyperparameter tuning to optimize this score. As a reference to compare our models to, we therefore use the \textit{baseline} that we trained ourselves.

%We introduce our own trained baseline which is used to compare the model scores.

On average, multi-head obtains slightly worse ROUGE scores. This can be a simple case of undertraining and sub-optimal choice of hyperparameters. However, we notice that in case of the Word Prior model, the multi-head architecture performs better. A possible explanation is that the decrease of weights in the pointer head might hurt the ROUGE score if the model mostly relies on the pointer, and when the generator gets a more important task the multi-head might be beneficial. This hypothesis needs more extensive training and tuning to prove, and is not in the scope of this research.

Dropout on the single head model achieves slightly higher ROUGE scores and does increase N-gram novelty. The multi-head dropout model is not significantly different from the baseline.

Both proposed losses greatly improve the number of novel N-grams and sentences. This is especially noticeable in case of the word prior loss; The number of novel N-grams is more than double. However, in both cases this increase in novel N-grams decreases the ROUGE score.

In example \ref{fig:example2}, we can clearly observe that the model favours generating over pointing when predicting simple words like articles or prepositions. On less common words, like names and uncommon nouns, the pointer is still used.

%%%%%%%%%%%%%%%%%%%
% - baseline: obtain lower 3 rouge points scores than original paper inline with observations with 
% - dropout:   minimal higher rouge-L over baseline, n-grams same as for baseline
% - Naive loss: rouge-L decreases by 1.5 compared to baseline, new n-grams 
% - WPLoss: same as for Naive loss but 
% - mh no difference to single head
% - vert correlates with rouge, other than claimed by the authors, no additional insight
%%%%%%%%%%%%%%%%%%%

\subsection{Is the model pointing less?} 

Table \ref{tab:pgen} shows the average $p_{gen}$ during train and test time, which show how much the model uses the pointer mechanism on average. The baseline has a $p_{gen}$ of 0.18 on average, which is in line with the findings of \citet{see2017get}. The Multihead does not change this behaviour. While the Dropout model uses the generator significantly more during training, during test time it falls back to the same value as the baseline.

Both loss functions greatly increase the amount the generator is used. This is to be expected: When actively penalizing the pointer mechanism, the pointer mechanism is used less at test time.

\begin{table}[!h]
\caption{Average value of $p_{gen}$ during the end of training and test time.}
\begin{tabular}{lll}
\midrule
Model         & train $p_{gen}$ & test $p_{gen}$ \\
\specialrule{0.01pt}{1pt}{0pt}
Baseline & 0.54            &  0.18              \\
4 Heads  & 0.51            &  0.17             \\
Dropout  & 0.67            & 0.17              \\
NLoss    & 0.77            & 0.32              \\
WPLoss   & 0.86            & 0.36            \\
\midrule
\end{tabular}
\label{tab:pgen}
\end{table}

The examples in Appendix B show the $p_{gen}$ value on each generated word for the baseline, and two new losses. The model trained word prior loss shows that it achieves a much higher average $p_{gen}$ on common words, such as articles and verbs. In the first example only three fragments have a low $p_{gen}$: "Ellie Meredith", "Down Syndrome", and "let". The first two fragments are cases where we want the model to point, whereas the third fragment is less clear. On inspection of the source article, this fragment starts a direct quote from the article: "Let's party like it's 1989".

\subsection{Novel Words}
% \todo[inline]{top new words}
To investigate the novel words that are used further the most frequent new words for the most abstractive model (multihead, coverage and WPLoss) is calculated (Tab. \ref{tab:new_words}). It can be noted that the majority of the words are verbs.
When a sentence is rephrased it can happen that the same root of a verb is used but with another suffix.
Similarly, when the tense of a verb changes this can introduce new words.
This suggests that the newly introduced words are valid rephrases and not random words. %This is maybe a stretch?
Note that the tokens < and > are the result of incorrect parsing of <unk> and <s> tokens.

\begin{table}[]
\caption{Most frequent novel words for the best model (4 heads, coverage and word prior loss).}
\label{tab:new_words}
\begin{tabular}{lllll}
\toprule
says   & beat   & diagnosed& unk& taken\\
scored & >   &say &premier& boss \\
been   &<    &:  &year & since\\
he     &has  & found & said &  by\\
\bottomrule
\end{tabular}
\end{table}

%\todo[inline]{Compare summary lengths}
The average reference summary length is 56 tokens. The length of the generated summaries is approximately 46 for models with a single attention head and word prior loss and around 55 for all other models (including the baseline). For none of the models the generated summary was on average longer than the target summaries. The maximum allowed length for generating summaries is 120. The new N-grams are thus not a result of simply generating more text because the average length does not change. Instead the novel N-grams replace non novel N-grams. This observation (and the type of newly generated words, as can be seen in table \ref{tab:new_words}) suggests that the novel N-grams are valid rephrasings and not random words or model artefacts.

 \section{discussion}

\subsection{Multi-head Evaluation}
% Although this differences have not been tested for significance most of the differences are bigger than $0.2$ which we found to be significant in the baseline comparison.
% ROUGE 2 WPLoss 14.93 (SH) -> 14.95. That is not significant
% but we can make our claim less strong by saing most of our measurements instead of all our measurements
The multi-head attention mechanism improves the results of the new loss function in our measurements. The ROUGE L score increases by $0.7$ (NLoss) and $0.6$ (WPLoss) but the novel N-grams drop slightly for both losses. This shows that penalizing the pointer mechanism when the attention between pointer and generator is shared can reduce the overall quality of summaries, which indicates the multi-head mechanism is working as intended: by splitting the pointer and generator attentions, penalizing the pointer affects the generator less.

\begin{figure}[H]
    \centering
    \includegraphics[width=0.7\linewidth]{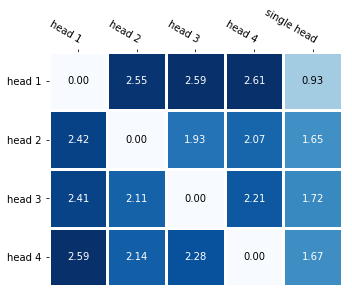}
    \caption{The average KL divergence between each pair of heads of the multi-head model, and between each head of the multi-head and the single head model (last column).}
    \label{fig:kld}
\end{figure}

Figure \ref{fig:kld} shows the KL divergence between each head of the multi-head attention, where cell (i, j) corresponds to $D_{KL}(head_i || head_j)$. In the last column the KL divergence between the attention distributions of the multi-head, and the attention distribution of the same model with just one attention head is shown. We can read from the plot that the head used for the pointer ('head 1') is on average very different from the other heads in the multi-head. This means that on average they attend to different words in the source text, and perform a different function when generating words. 

On the other hand, the pointer head is most similar to the attention distribution in the single head model. Additionally, the other heads in the multi-head are more similar to the single head than to eachother. This result indicates that the single head model attempts to incorporate information needed for both pointing and generating, but that it can be desirable to split this information into multiple heads.

\subsection{Dropout}
Models that used the dropout mechanism show an increased ROUGE score while having lowered amounts of novel N-grams. This is the exact opposite of what was expected.
A possible explanation is that dropout does not produce gradients that indicate that the pointing mechanism is wrong but that it only slows its training. Still, it does force the generator to be more reliable and this leads to better scores in general. This is in line with the idea of \citet{see2017get} that the generator might not be optimally trained because of over-reliance on the pointer. Using dropout is therefore very similar to the baseline model expect that dropout does give the generator the opportunity to be better trained.

\subsection{Coverage ablation studies}
\label{sec:ablation}
To understand the relationship between the coverage loss and our proposed losses we trained all models without the coverage mechanism. Tables \ref{rouge_a} and \ref{ngram_a} in the appendix show that the Rouge-L score decreases about 2 points for all models without using the coverage loss. Further, it can be seen that the number of novel N-grams decreases significantly. This matches our expectations for N-grams > 1 which reflect the order of words. However, we can also observe a decrease in novel 1-grams which suggests that the coverage mechanism favours extractive summarization which stands in contrast to our goal towards a more abstractive model. This claim is supported by table \ref{tab:dup_ngram} which shows that introducing the coverage mechanism reduces the amount of repetition whereas both WPLoss and NLoss result in more duplicate N-grams. The losses thus interfere with coverage. Coverage reduces repetition at the expense of more extractive summaries. Penalizing the pointing mechanism introduces new N-grams but also increases the problem of overgeneration.

\begin{table}[]
\caption{Duplicated N-grams within summaries for multi-head models where cov stands for models that have been trained with coverage.}
\label{tab:dup_ngram}
\resizebox{\columnwidth}{!}{%
\begin{tabular}{llllllll}
\toprule
Model extensions &    & 1-grams & 2-grams & 3-grams & 4-grams & Sentences \\
\midrule
\textit{baseline} & &  0.32  &  0.20  &  0.19  & 0.18   &  0.02       \\ 
 NLoss &       &   0.34 &   0.24 &  0.22  & 0.20   &   0.03    \\ 
 WPLoss &        & 0.40 & 0.29   & 0.27   & 0.25   & 0.00   \\ 
\textit{baseline} &    cov & 0.22   & 0.07   & 0.06 & 0.05   &  0.00     \\ 
 NLoss &      cov    &  0.24  &  0.11  &  0.09  & 0.08  &    0.01   \\ 
 WPLoss &      cov   & 0.29    & 0.16   & 0.13    & 0.12    & 0.01     \\ 
\bottomrule
\end{tabular}
}%
\label{tab:dublicated_ngrams}
\end{table}

\subsection{Pointer and generator distributions}
When examining the examples produced by the model, we notice that even with a high $p_{gen}$, the model is copying full sentences. This effect can be seen in Figure \ref{fig:example4} (Appendix B): The last sentence is mostly made with the generator, but is a copy of line 11 in the source article.
An issue of the pointing distribution is that it has a much lower cardinality compared to the generator distribution, in our case it is two orders of magnitude smaller (401 for the pointer, 50k for the generator). This means that there is generally a bias towards words from the source text, even with high values of $p_{gen}$. We can conclude that the pointer mechanism by definition decreases the novel 1-gram score, and makes the model less abstractive. 

Instead of learning the soft switch $p_{gen}$, which introduces this bias, a hard pointing mechanism could be learnt by reinforcement learning or with a Gumbel-Softmax approximation \cite{jang2016categorical} to diminish this bias.

\subsection{ROUGE metric and dataset}
The aforementioned problems of ROUGE are a serious limitation to the evaluation of abstractive summarization models. The idea of abstractive summarization is that valid summaries can be created in many ways. However, ROUGE measures the exact overlap between a set of references summaries and summaries created by our models. This means that either the set of reference summaries should be representative of all valid ways in which abstractive summaries can be produced or that there is a need for a new metric that does not rely on exact overlap but rather on semantic similarity. 

\section{conclusion}
In this work, we investigated several additions to the Pointer-Generator framework in order to improve abstraction while maintaining similar summary quality. 

While the multi-head mechanism learns different features for pointing and generating, it does not improve the ROUGE score. When dropout is used on the pointer mechanism, the multi-head attention does promote novel N-grams in the produced summary, but in other models the results are very similar.
% \todo[inline]{Conclusion: Multihead does not help ROUGE, does help novel ngrams with new losses.}

% \todo[inline]{Conclusion: Dropout seems to help ROUGE score, but not novel ngrams}
The dropout mechanism does the opposite of what was expected as the amount of novel N-grams decreased while the ROUGE scores increased. It seems likely that dropout partly removes the over-reliance on the pointer and therefore gives improved performance compared to the baseline.
The two introduced loss functions improve the generation of novel N-grams significantly. For the Word Prior loss, we observe an improvement of 12.5\% more novel sentences compared to the baseline model. However, in both cases we did not manage to maintain the same ROUGE scores. This might be a problem with the loss functions, but also with the training process.

The VERT metric looked like a promising metric to measure the semantic similarity between generated and target summaries. However, the resulting VERT scores are completely correlated with the ROUGE metric and do not produce any new insights.

As our goal was to train a model that is more abstractive we used the number of novel N-grams with respect to the reference summary as a measure. The novel N-grams score can easily be increased by adding random words to the summary. This does not lead to more abstractive or higher quality summaries. Since there is no metric to evaluate abstractive summarization tasks effectively it is not possible to claim that the summaries are more abstractive based on just the increase of new N-grams. However we have shown that the summary length does not increase which suggests that new words replace other words instead of adding more words. We have also shown that the novel words are plausible words for rephrasing/summarization tasks.

\subsection{Future work}
It appears that the newly introduced losses interfere with the coverage mechanism and increase the over-generation problem. It could be the case that by introducing the coverage and new loss at the same point in training produces gradients that are too different from training without these losses, which would interfere with the convergence of the network. In future work, the interaction between these new loss function and the coverage can be investigated.

The difference in pointing behaviour between training and inference could be reduced by using scheduled teacher-forcing \cite{bengio2015scheduled} , which gradually decreases the frequency the model receives the ground truth as input to the generator. This reduces the difference between training and inference, which could result in higher values of $p_{gen}$.

\nocite{*}
\bibliographystyle{ACM-Reference-Format}
\bibliography{bibliography}

\newpage
\begin{appendices}
\section{Scores without coverage}\label{sec:appA}

\begin{table}[!h]
\caption{Mean rouge $\boldsymbol{F_{1}}$ and VERT scores of the tested models (11k examples in testset). All models were trained without coverage. Here NLoss corresponds to naive pointing loss and WPLoss to the word prior pointing loss.}
\resizebox{\columnwidth}{!}{%
\begin{tabular}{lllllllll}
\toprule
%Model                 & Rouge 1 & Rouge 2 & Rouge L & VERT  \\ 
%Model & & & & & & & & \\
Attention heads & \multicolumn{2}{l}{Model extensions} & Rouge 1 & Rouge 2 & Rouge L & VERT  \\ 
%MH & cov & Dropout & NLoss & WPLoss
\midrule
1 &   &           & 36.65   & 15.30   & 31.53   & 0.691 \\ % 
4 &   &              & 36.50   & 15.15   & 31.43   & 0.691    \\ % jorn
\specialrule{0.01pt}{1pt}{0pt}
 1&  dropout &           & 36.37 & 15.20 & 31.13 & 0.688 \\% david 
 4&   dropout &         &36.27 & 15.11&31.19 & 0.689 \\% david
 \specialrule{0.01pt}{1pt}{0pt}
 %WPLoss (.05) + cov & 37.47   & 15.61   & 32.79   & 0.699\\ % freek
 %MH + WPLoss (.05) + cov  & 37.57   &  15.52  & 33.02   & 0.701  \\ % david
 %MH + WPLoss (.05) + cov  ( 2150 steps)& 36.73   &  15.17  & 31.87   & 0.694  \\ % david
 %WPLoss (.10) + cov        & 36.95       & 15.25        & 32.16      & 0.695     \\ % jorn 0.1
 %MH + WPLoss (.10) + cov  & 37.37   &  15.41  & 32.75   & 0.698  \\ % david
 1&   &NLoss   &  35.87 & 14.96 & 30.81  & 0.692 \\
 4&   &  NLoss  &   35.55&  14.58  &   30.37 & 0.683 \\
\specialrule{0.01pt}{1pt}{0pt}
 1&   &   WPLoss  &35.44   & 14.49   & 30.13  & 0.680\\
 4 &   &   WPLoss & 35.12 & 14.26 & 30.01   & 0.678   \\
 %MH + WPLoss (.10) + cov (7408 steps) & 36.74   &    15.14     &    32.04     &    0.694   \\  % david
 %2H + WPLoss(.2) + cov & 36.83 & 14.99 & 32.12 & 0.691 \\
 %2H + WPLoss(.05) + cov & 37.031 & 14.91 & 32.627 & 0.697 \\
\specialrule{0.8pt}{1pt}{0pt}
1 (\textit{baseline \citet{see2017get}})&  & &  36.44& 15.66 & 34.42 & -\\
\bottomrule
\end{tabular}
}%
\label{rouge_a}
\end{table}

\begin{table}[h!]
\caption{Percentage of novel N-grams and sentences that are produced for each of the tested models.}
\resizebox{\columnwidth}{!}{%
\begin{tabular}{llllllllll}
\toprule
Attention heads & \multicolumn{2}{l}{Model extensions}         & 1-grams & 2-grams & 3-grams & 4-grams & Sentences \\
\midrule
1  & &           & 0.36    & 4.31    & 10.51   & 16.25   & 82.50     \\ 
4 & &     & 0.31    & 3.97    & 9.84    & 15.41   & 81.40      \\
\specialrule{0.01pt}{1pt}{0pt}
1 & Dropout&           & 0.28    & 4.28    & 10.74   & 16.80   & 83.31      \\ 
4 &Dropout &      & 0.60     & 4.37    & 10.42   & 16.08   & 82.66      \\ 
\specialrule{0.01pt}{1pt}{0pt}
%WPLoss (.05) + cov     &   0.36      & 4.92     & 11.25   & 17.05   & 85.62                   \\ 
%WPLoss (.05) + MH + cov (2150 steps) &  0.39   &  5.26   &    11.92     &     17.95    &    88.26       \\
%WPLoss (.05) + MH + cov  &  0.23   &  4.64   &    11.14     &     17.15    &    84.72       \\
%WPLoss (.10) + cov       & 0.64    &  6.08   &    13.38     &     19.89    &    88.05       \\ 
%WPLoss (.10) + MH + cov  &  0.37   &  5.58   &    13.11     &     19.86    &    87.13       \\
1 & &NLoss     & 0.28 & 5.11 & 12.68 & 19.29 & 86.54\\
4 & &NLoss     & 0.31 &5.28  & 12.87 & 19.73 & 87.22 \\
\specialrule{0.01pt}{1pt}{0pt}
1 & &WPLoss       & 1.43 & 8.90 & 18.49 & 26.46 & 92.19 \\
4 & &WPLoss       & 0.90 & 8.19 & 18.40 & 26.97 & 91.35 \\
% WPLoss (.10) + MH + cov (7408 steps) &    0.65     &     6.53    &     14.40    &      21.33   &        88.13   \\
% 2H + WPLoss(.20) + cov & 0.55 & 7.93 & 16.94 & 25.92 & 90.94 \\
% 2H + WPLoss(.05) + cov & 0.41 & 5.01 & 11.41 & 17.64 & 84.79 \\
\specialrule{0.8pt}{1pt}{0pt}
\multicolumn{3}{l}{Target summaries} & 16.95 & 52.48 & 72.36    & 81.94 & 98.97 \\
\bottomrule
\end{tabular}
}%
\label{ngram_a}
\end{table}
\onecolumn
\section{Examples}
The following pages contain randomly chosen examples from the Multihead model with coverage, combined with the new losses. 
%Due to a visualization problem, out of vocabulary words are not shown in the new summaries. 
The words highlighted in red reflect the overall attention the model paid to a word during the constructing the summary. Italic words denote out-of-vocabulary words. The green shading intensity represents the value of the generation probability $p_gen$.

\begin{figure}[h!]
    \centering
    \includegraphics[width=\textwidth]{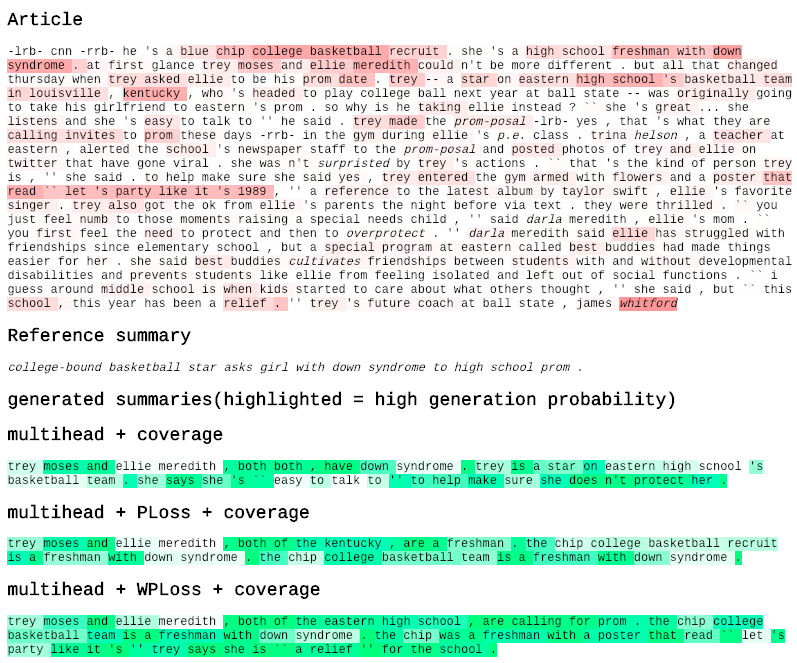}
    \centering
    \caption{In this example, the average $p_{gen}$ is much higher in the word prior loss model, except on words with a low prior ('ellie meredith', 'down syndrome').}
    \label{fig:example1}
\end{figure}

\begin{figure}[h!]
    \centering
    \includegraphics[width=\textwidth]{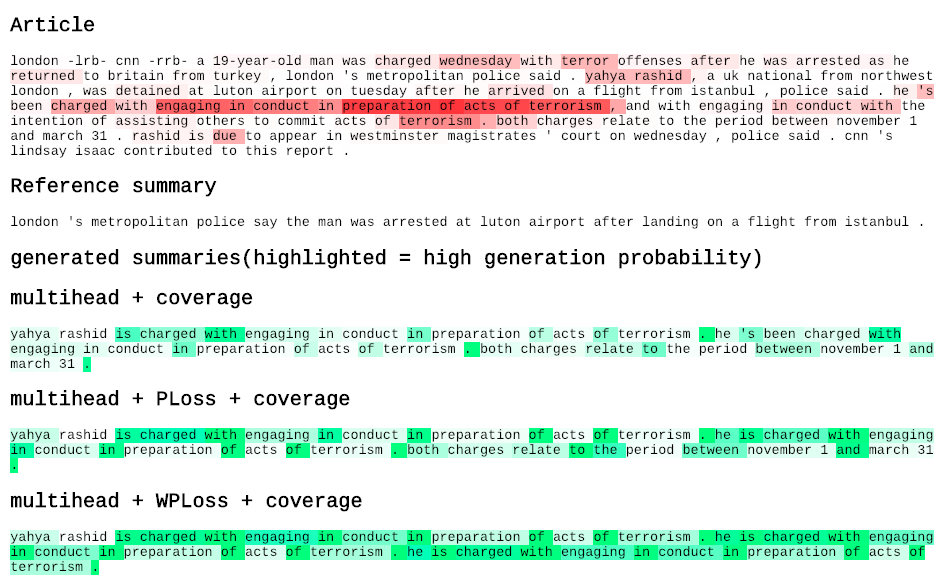}
    \centering
    \caption{This example shows clearly that the new losses re-introduce the over generation problems that the coverage loss aimed  to solve.}
    \label{fig:example2}
\end{figure}
\begin{figure}[h!]
    \centering
    \includegraphics[width=\textwidth]{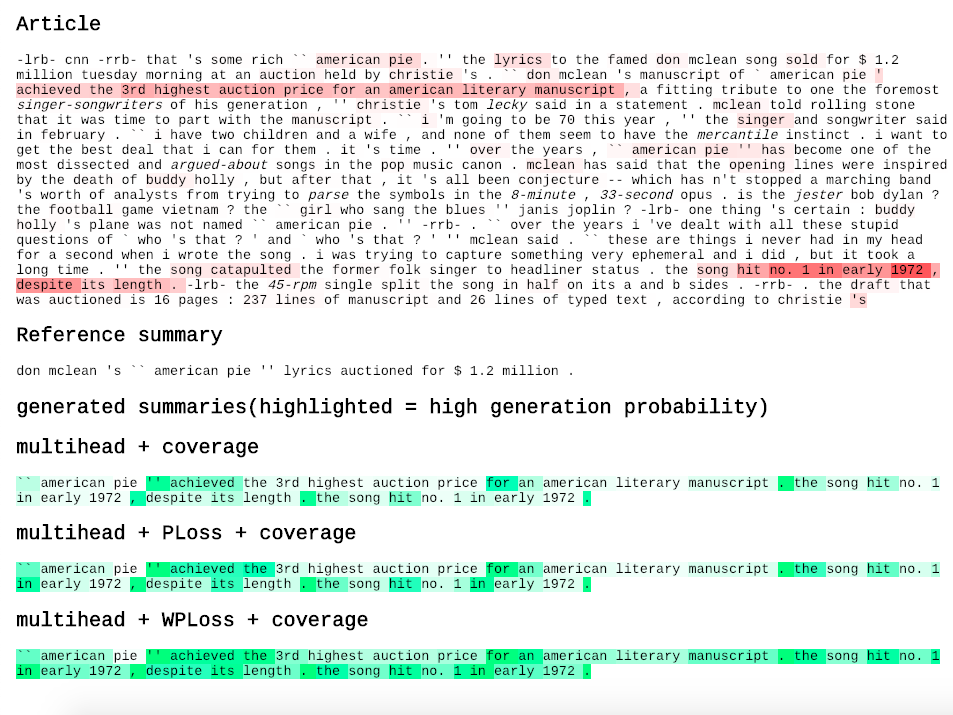}
    \centering
    \caption{Another example. While the average $p_{gen}$ of each word is significantly higher in the WPLoss model, the generated summaries are the same.}
    \label{fig:example3}
\end{figure}

\begin{figure}[h!]
    \centering
    \includegraphics[width=\textwidth]{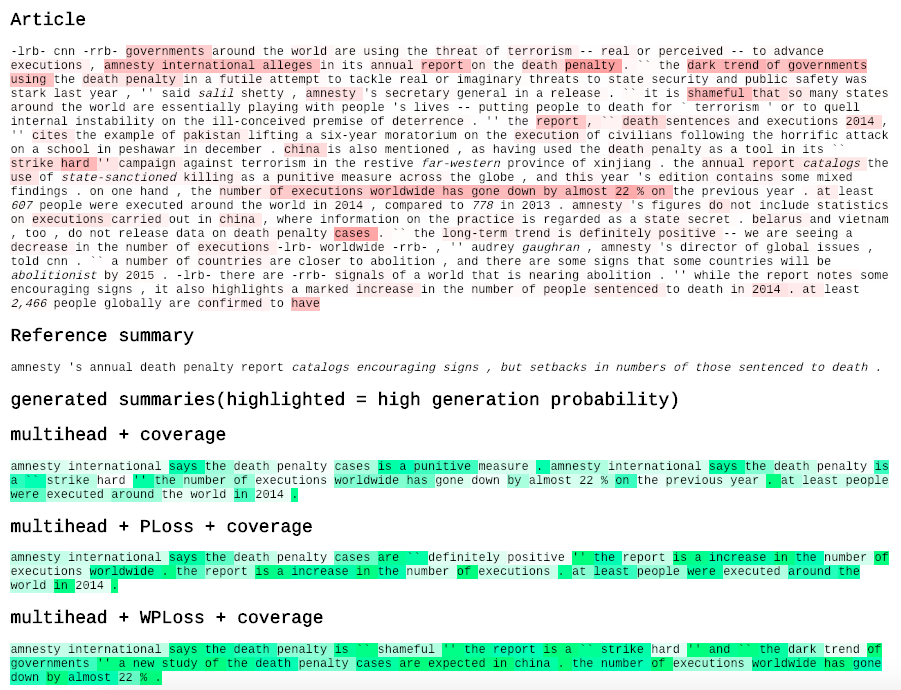}
    \centering
    \caption{The last sentence in the WPLoss example on average mostly uses the generator, but is an exact copy of line 11 in the source article.}
    \label{fig:example4}
\end{figure}

\end{appendices}

\end{document}